\documentclass[review,12pt]{elsarticle}

\usepackage[utf8]{inputenc}
\usepackage[T1]{fontenc}
\usepackage{amsmath,amssymb}
\usepackage{graphicx}
\usepackage{booktabs}
\usepackage{multirow}
\usepackage{array}
\usepackage{enumitem}
\usepackage{xcolor}
\usepackage{url}
\usepackage{hyperref}

\biboptions{sort&compress}

\newcommand{\RegOnly}{RankR-NS-RegOnly}
\newcommand{\FullNS}{RankR-NS}
\newcommand{\MacroF}{Macro-F1}

\journal{Knowledge-Based Systems}

\begin{document}

\begin{frontmatter}

\title{Prototype-Rule Neurosymbolic Regularization for Rank-Constrained Tensor Neural Networks under Label Scarcity}

\author[uom]{Eftychios Protopapadakis\corref{cor1}}
\ead{eftprot@uom.edu.gr}
\author[umalta]{Konstantinos Makantasis}
\author[uom]{Konstantinos M. Giannoutakis}

\cortext[cor1]{Corresponding author.}
\address[uom]{Department of Applied Informatics, University of Macedonia, 156 Egnatia Street, 54636 Thessaloniki, Greece}
\address[umalta]{Department of Artificial Intelligence, Faculty of Information \& Communication Technology, University of Malta, Msida, Malta}

\begin{abstract}
Rank - constrained tensor neural networks reduce the parameterization of high - order inputs, but they do not explicitly constrain class geometry in the learned representation. This study investigates whether a differentiable prototype - rule can provide a complementary inductive bias for Rank-R tensor learning under limited supervision. The proposed framework augments the Rank-R objective with prototype-based regularization and optionally fuses prototype evidence with neural logits at inference. Four hyperspectral benchmarks are evaluated with four Rank-R configurations under both seven-fold stratification and spatially separated folds that mitigate leakage; a separate spatial study varies the class support budget from 2 to 20 samples. Under spatial evaluation, full neurosymbolic inference changes Macro-F1 score by +8.82 percentage points on Botswana, +5.49 on Indian Pines, +1.59 on Pavia University, and -0.62 on Salinas. Most of the benefit arises from training time regularization, whereas inference fusion is small and dataset dependent.
\end{abstract}


\end{frontmatter}


\section{Introduction}
\label{sec:introduction}


Tensor-valued observations arise naturally in remote sensing and in other high-dimensional problems, in which the ordering of input modes carries information. A conventional dense neural layer can ignore that structure and may introduce many parameters relative to the number of labeled observations. Rank-constrained feedforward neural networks \cite{makantasis2018tensor,makantasis2019hyperspectral} address this issue by representing the input-to-hidden weight tensors through a Canonical/Polyadic (CP) decomposition rather than learning unrestricted dense tensors \cite{makantasis2021rankr,kolda2009tensor}. The resulting Rank-R formulation retains multilinear structure while substantially reducing the number of free parameters, an attractive property for small-sample high-order learning \cite{makantasis2021rankr}. Rank-R representations have also been reused as embeddings for graph-based semi-supervised hyperspectral classification \cite{georgoulas2023graph}.

Nevertheless, parameter efficiency constrains how the mapping is represented, rather than how classes should be arranged in the learned latent space. A complementary line of research incorporates prior relations or logical constraints into differentiable learning objectives \cite{roychowdhury2021regularizing,badreddine2022ltn,hitzler2022neurosymbolic,li2023neuro}. Prototype-based methods provide a particularly simple class-level structure: examples of a class are related to a representative point in embedding space, and membership can be expressed through relative compatibility with class prototypes \cite{snell2017prototypical,ding2024integrating}. This suggests a natural question for Rank-R learning: can an explicit prototype relation provide information that is complementary to the low-rank tensor constraint?

The present work answers that question as a controlled characterization, rather than as a state-of-the-art hyperspectral-classification benchmark. We introduce a differentiable prototype-rule regularizer into a Rank-R neural model and deliberately separate two possible mechanisms: representation shaping during training and the use of prototype evidence at inference. We further distinguish conventional stratified evaluation from spatially separated evaluation because patch-based hyperspectral experiments can be optimistic when neighboring samples from the same scene are allowed to cross split boundaries \cite{feng2023leakage,cao2023disjoint}. 

Finally, we evaluate the mechanism under explicit per-class label budgets to test the common intuition that stronger inductive biases should become more useful as supervision decreases. This hypothesis is consistent with prior work in constraint-based prior-knowledge regularization and prototype-based few-shot learning, where additional structural information is introduced specifically to improve learning when direct supervision is limited \cite{roychowdhury2021regularizing,snell2017prototypical}.

To the best of our knowledge, prior Rank-R FNN studies have not investigated explicit differentiable class rules as an additional training constraint. The novelty claim is therefore intentionally narrow: we study prototype-rule neurosymbolic regularization of CP-constrained Rank-R representations and characterize when its contribution is positive, neutral, or adverse.

\subsection{Contributions and research questions}
\label{subsec:contributions}

The paper makes four contributions. First, it integrates a differentiable prototype-rule mechanism into a CP-constrained Rank-R neural representation. Second, it introduces a matched ablation that isolates training-time prototype regularization from inference - time prototype fusion without changing the Rank-R backbone. Third, it evaluates the same mechanism under both spatially leakage-controlled and conventional stratified protocols and keeps their evidence separate. Fourth, it performs a dedicated spatial label-scarcity study over seven per-class budgets and five support realizations, allowing the dependence of the neurosymbolic effect on supervision, dataset, and Rank-R configuration to be examined directly. 

These contributions are organized around five research questions. The first concerns whether the mechanism works at all: does differentiable prototype - rule regularization change classification performance relative to the corresponding purely neural Rank-R baseline (RQ1)? Where an effect occurs, we ask how much of it is attributable to training - time regularization and how much to prototype-based inference fusion (RQ2), and how the observed effect differs between conventional stratified evaluation and spatially leakage - controlled evaluation (RQ3). We further ask whether the contribution of the prototype-rule mechanism systematically increases as the number of labeled samples per class decreases (RQ4), and to what extent the effect is conditioned by dataset and Rank-R architecture (RQ5).

\section{Related Work}
\label{sec:related}
The proposed framework draws on two lines of work that have largely been developed separately. Rank-constrained tensor networks address the parameterization of high-order inputs but leave the geometry of the learned representation unconstrained, while neurosymbolic and prototype-based methods impose explicit class-level structure without regard on how the underlying representation is parameterized. Section \ref{subsec:related-rankr} reviews the former, Section \ref{subsec:related-ns} the latter, and Section \ref{subsec:positioning} positions the present work as a deliberately minimal combination of the two.

\subsection{Rank-R tensor neural networks}
\label{subsec:related-rankr}

The Rank-R FNN constrains each input-to-hidden weight tensor through a CP decomposition, preserving the high-order organization of the input while replacing a full tensor by a sum of rank-one factors \cite{makantasis2021rankr,kolda2009tensor}. This parameterization is especially relevant when the input dimensionality is high but labeled data are limited. Subsequent work has reused Rank-R-derived tensor embeddings in graph-based semi-supervised hyperspectral classification, showing that the representation can support additional relational structure beyond the original classifier \cite{georgoulas2023graph}. The present work differs from these studies by changing the learning objective itself through an explicit class-level constraint.

\subsection{Neurosymbolic and prototype-based learning}
\label{subsec:related-ns}

Neurosymbolic learning broadly combines neural representation learning with symbolic, logical, or knowledge-based structure \cite{hitzler2022neurosymbolic,wang2025}. Differentiable approaches such as Logic Tensor Networks and constraint-based regularization map relations into continuous objectives that can participate in gradient-based optimization \cite{badreddine2022ltn,roychowdhury2021regularizing,li2023neuro}. The symbolic component need not be a discrete theorem prover; it can be a declarative relation whose degree of satisfaction is made differentiable \cite{marra2024}. In the present framework, this means that the symbolic knowledge is the explicit class-level rule that an embedding should be closer to the prototype of its own class than to competing class prototypes; the degree to which this rule is satisfied is converted into a differentiable loss and optimized jointly with the neural model.

Prototype-based learning offers a related geometric mechanism. Prototypical networks, for example, classify a query according to its relation to class prototypes computed in an embedding space \cite{snell2017prototypical,zhang2024weighted}. Prototype learning has also been explored in hyperspectral image classification under limited labeled supervision \cite{ding2024integrating}, including prototype-rectification strategies designed specifically for cross-domain few-shot classification \cite{Qin2024}. More recent work has further developed global–local prototype representations to construct discriminative spatial–spectral metric spaces for few-shot HSI classification \cite{tang2025}.

In the present study, prototypes are not used as a meta-learning episode construction. Instead, they ground the explicit relation that an embedding belonging to class $c$ should be more compatible with prototype $c$ than with competing class prototypes. The resulting term is used as a training constraint, while prototype-based inference is treated as a separable optional component.

\subsection{Positioning of the present work}
\label{subsec:positioning}

The proposed model is deliberately a minimal neurosymbolic construction.  Unlike approaches employing knowledge-graph reasoning \cite{delong2024neurosymbolic}, ontology-based knowledge representation \cite{armary2025ontology}, or
symbolic state-space search \cite{fivser2024boosting},
the proposed framework relies on a differentiable
class-level relation. This relation explicitly constrains
the geometry of the learned representation.
The resulting formulation enables a controlled evaluation
of the additional inductive bias while preserving
the underlying Rank-R architecture.


\begin{table}[t]
\centering
\caption{Conceptual positioning of the proposed mechanism. The table compares mechanisms rather than predictive performance.}
\label{tab:positioning}
\small
\resizebox{\linewidth}{!}{%
\begin{tabular}{lcccc}
\toprule
Approach & Tensor structure & Explicit relation & Training constraint & Prototype inference \\
\midrule
Rank-R FNN & \checkmark & -- & -- & -- \\
Prototype-based learning & varies & -- & varies & varies \\
Differentiable constraint-based NS & varies & \checkmark & \checkmark & varies \\
\textbf{Present framework} & \checkmark & \checkmark & \checkmark & optional \\
\bottomrule
\end{tabular}}
\end{table}

\section{Proposed Neurosymbolic Rank-R Framework}
\label{sec:framework}
\begin{figure}[t]
\centering
\includegraphics[width=0.88\linewidth]{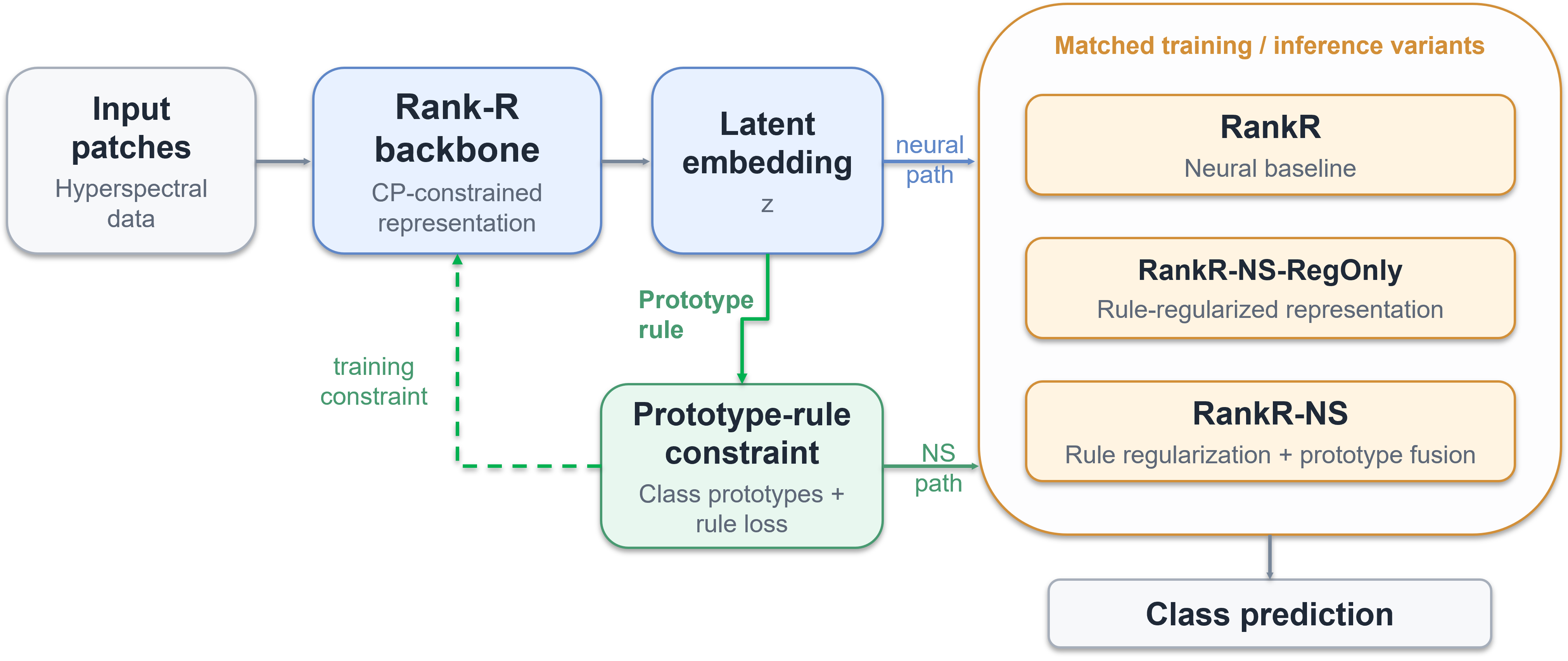}
\caption{High-level overview of the proposed neurosymbolic Rank-R framework. Hyperspectral input patches are mapped to a latent representation through the CP-constrained Rank-R backbone. The learned embedding can be further shaped by the prototype-rule constraint, which introduces class-prototype attraction and inter-class separation during training. The resulting representation supports three matched variants: the purely neural RankR baseline, RankR-NS-RegOnly using prototype-rule regularization during training, and RankR-NS additionally incorporating prototype evidence during inference.}
\label{fig:flowchart}
\end{figure}

This section defines the proposed framework in three parts. Section \ref{subsec:backbone} recalls the CP-constrained Rank-R backbone that produces the latent embedding used throughout. Section \ref{subsec:rule} introduces the differentiable prototype-rule regularizer, its class-prototype distance, and its combination with the cross-entropy objective. Section \ref{subsec:variants} then defines the three matched training and inference variants — RankR, RankR - NS - RegOnly, and RankR-NS — used to separate training-time regularization from inference-time prototype fusion. Figure \ref{fig:flowchart} provides a high-level overview of the proposed neurosymbolic Rank-R framework.

\subsection{Rank-R backbone and latent representation}
\label{subsec:backbone}

Let an input sample be a $D$-order tensor $\mathcal{X}\in\mathbb{R}^{I_1\times\cdots\times I_D}$. For hidden unit $q$, the dense weight tensor is replaced by a rank-$R$ CP representation,
\begin{equation}
\mathcal{W}^{(q)} = \sum_{r=1}^{R}
\mathbf{w}_{1}^{(q,r)} \circ \mathbf{w}_{2}^{(q,r)} \circ \cdots \circ \mathbf{w}_{D}^{(q,r)},
\label{eq:cp}
\end{equation}
where $\circ$ denotes the outer product. he output $z^{(q)}$ of hidden unit $q$ is the computed as 
\begin{equation}
    z^{(q)} = f(\text{vec}(\mathcal{X})^T \cdot \text{vec}(\mathcal{W}^{(q)})),
\end{equation}
where $\text{vec()}$ is the vectorization operator which transforms tensor into a column vector, and $f()$ is a non-linear activation function. With $H$ neurons in the first hidden layer, the input tensor is transformed into a vector embedding 
\begin{equation}
    \mathbf{z} = [z^{(1)} z^{(2)} \cdots z^{(H)}] \in \mathbb{R}^H.
\end{equation}
The ordinary Rank-R classifier maps $\mathbf{z}$ to neural class logits and is optimized by cross-entropy.

\subsection{Prototype-rule neurosymbolic regularization}
\label{subsec:rule}

Each class $c$ is associated with a prototype vector $\mathbf{p}_c\in\mathbb{R}^{H}$. The prototypes are inactive during an initial cross-entropy warm-up. At rule activation they are initialized from the mean embedding of the labeled training samples of the corresponding class,
\begin{equation}
\mathbf{p}_c \leftarrow
\frac{1}{|\mathcal{S}_c|}
\sum_{i\in\mathcal{S}_c}\mathbf{z}_i ,
\label{eq:prototype}
\end{equation}
where $\mathcal{S}_c$ contains only training samples with label $c$. This mean is an initialization rather than a permanent recomputation: after activation, $\mathbf{p}_c$ remains a trainable model parameter and is updated jointly with the Rank-R factors and neural classifier by backpropagation.

The explicit class-level relation represented by the
rule can be stated as follows: IF $y_i=c$ THEN $\mathbf{z}_i \text{ should be close to } \mathbf{p}_c$ and farther
from competing class prototypes. 
The implementation grounds this relation with cosine distance. Defining normalized embeddings and prototypes as $\bar{\mathbf{z}}_i=\mathbf{z}_i/\|\mathbf{z}_i\|_2$ and $\bar{\mathbf{p}}_c=\mathbf{p}_c/\|\mathbf{p}_c\|_2$, the class-prototype distance is
\begin{equation}
d_{ic}=1-\bar{\mathbf{z}}_i^{\top}\bar{\mathbf{p}}_c .
\label{eq:proto-distance}
\end{equation}
For sample $i$, let $d_i^{+}=d_{iy_i}$ and let
\begin{equation}
d_i^{-}=\min_{c\neq y_i} d_{ic}
\end{equation}
denote the distance to the closest competing prototype.The implemented truth value of the rule for the labeled class is
\begin{equation}
t_i=\max\!\left(\exp\!\left[-d_i^{+}/\tau\right],\,10^{-8}\right),
\label{eq:rule-truth}
\end{equation}
and the minibatch rule loss is
\begin{equation}
\mathcal{L}_{\mathrm{rule}}
=
-\frac{1}{B}\sum_{i=1}^{B}\log t_i
+
\frac{1}{B}\sum_{i=1}^{B}
\max\!\left(0,\,m+d_i^{+}-d_i^{-}\right).
\label{eq:rule-loss}
\end{equation}
Thus the first term attracts an embedding toward its labeled-class prototype, while the second requires the nearest competing prototype to be at least margin $m$ farther away. After warm-up, this constraint is combined with the ordinary neural cross-entropy,
\begin{equation}
\mathcal{L}_{\mathrm{total}}
=
\mathcal{L}_{\mathrm{CE}}
+
\lambda_{\mathrm{rule}}\mathcal{L}_{\mathrm{rule}} .
\label{eq:total-loss}
\end{equation}
Across all neurosymbolic experiments,$\lambda_{\mathrm{rule}}=0.1$, $m=0.1$, and $\tau=0.25$. The rule activates after 20 warm-up epochs in the standard experiments and after 60 optimization updates in the fixed-update label-scarcity study. Because the prototypes are registered model parameters before the optimizer is created, gradient updates after activation jointly modify the prototypes and the neural representation. The reported experiments use a single fixed setting for these rule hyperparameters across every dataset, split protocol, and Rank-R architecture. No dataset-specific retuning is applied, so the reported comparisons reflect the same rule strength, margin, and temperature throughout the study rather than per-dataset optimization.

\subsection{Training and inference variants}
\label{subsec:variants}

Three matched variants isolate the two proposed mechanisms. \textbf{RankR} is the purely neural baseline trained with cross-entropy. The \textbf{regularization - only variant} (\RegOnly{}) uses the neurosymbolically trained checkpoint but evaluates it with the ordinary neural classifier; its contrast with RankR measures training - time representation shaping. The \textbf{full NS variant} (\FullNS{}) uses the same checkpoint and adds prototype evidence at inference. Specifically, the prototype score for class $c$ is
\begin{equation}
r_{ic}=-\frac{d_{ic}}{\tau},
\label{eq:rule-logit}
\end{equation}
and, if $\ell_{ic}$ is the neural class logit, the fused logit is
\begin{equation}
\tilde{\ell}_{ic}
=
\ell_{ic}
+
\beta r_{ic}
=
\ell_{ic}
-
\beta\frac{d_{ic}}{\tau}.
\label{eq:fusion}
\end{equation}
The experiments use $\beta=0$ for \RegOnly{} and $\beta=0.5$ for \FullNS{}. Consequently, both variants use the same neurosymbolically trained parameters and prototypes, and differ only in whether the prototype-derived logits alter the final decision. The shared checkpoint prevents the fusion comparison from being confounded by a second training run.

\begin{table}[t]
\centering
\caption{Experimental variants and mechanistic interpretation.}
\label{tab:variants}
\small
\resizebox{\linewidth}{!}{%
\begin{tabular}{lcccc}
\toprule
Method & CE & Rule regularization & Prototype fusion & Scientific role \\
\midrule
RankR & \checkmark & -- & -- & Purely neural Rank-R baseline \\
\RegOnly & \checkmark & \checkmark & -- & Training-time rule effect \\
\FullNS & \checkmark & \checkmark & \checkmark & Regularization plus inference fusion \\
\bottomrule
\end{tabular}}
\end{table}

\section{Experimental Methodology}
\label{sec:methodology}
This section describes the experimental setup used to answer the research questions in Section \ref{subsec:contributions}. Section \ref{subsec:datasets} introduces the four hyperspectral datasets and the input construction; Section \ref{subsec:training} specifies the Rank-R architecture grid and training controls. Section \ref{subsec:protocols} defines the spatial leakage-controlled and stratified evaluation protocols, and Section \ref{subsec:scarcity} details the label-scarcity protocol built on the spatial split. Section \ref{subsec:stats} closes with the Macro-F1-based evaluation and the statistical procedures used to compare methods.

\subsection{Datasets and input construction}
\label{subsec:datasets}

Experiments use four established hyperspectral scenes: Pavia University, Indian Pines, Salinas, and Botswana\footnote{Grupo de Inteligencia Computacional, University of the Basque Country (UPV/EHU), Hyperspectral remote sensing scenes, Public hyperspectral benchmark repository, accessed September 19, 2026. URL: \url{https://www.ehu.eus/ccwintco/index.php?title=Hyperspectral_Remote_Sensing_Scenes}}. Their standard corrected forms differ substantially in spatial extent, number of classes, and labeled sample count (Table ~\ref{tab:datasets}), providing both urban and vegetation-dominated cases. Each labeled center pixel is represented by a $5\times5$ spatial patch retaining the available spectral channels.Each extracted patch is independently normalized to the $[0,1]$ range using min--max scaling computed over all spatial and spectral entries within that patch.

The study uses 103 bands for Pavia University, 200 for Indian Pines, 204 for Salinas, and 145 for Botswana, consistent with the commonly used corrected versions in which unusable/noisy bands have already been removed where applicable. The $5\times5$ neighborhood provides local spatial context while remaining compact enough to support the dead-zone-constrained spatial splitting strategy used below; larger neighborhoods would increase the area that must be excluded around split boundaries and would further reduce the number of feasible spatial partitions.

\begin{table}[t]
\centering
\caption{Hyperspectral datasets used in the experiments. ``Bands'' denotes the usable channels in the corrected data used by the experiments.}
\label{tab:datasets}
\small
\begin{tabular}{lrrrrr}
\toprule
Dataset & Spatial size & Bands & Classes & Labeled pixels & Patch \\
\midrule
Pavia University & $610\times340$ & 103 & 9 & 42,776 & $5\times5$ \\
Indian Pines & $145\times145$ & 200 & 16 & 10,249 & $5\times5$ \\
Salinas & $512\times217$ & 204 & 16 & 54,129 & $5\times5$ \\
Botswana & $1476\times256$ & 145 & 14 & 3,248 & $5\times5$ \\
\bottomrule
\end{tabular}
\end{table}

\subsection{Architecture grid and training controls}
\label{subsec:training}

The Rank-R grid is the Cartesian product $R\in\{3,5\}$ and hidden dimension $H\in\{50,75\}$, yielding R3-H50, R3-H75, R5-H50, and R5-H75. These configurations are fixed experimental conditions rather than statistical replicates. All standard runs use model seed 1, AdamW with learning rate $2\times10^{-3}$ and weight decay $10^{-4}$, batch size 64, gradient-norm clipping with a maximum norm of 5, and at most 50 training epochs, subject to early stopping. 

Model selection uses validation \MacroF{} from the ordinary neural logits, including for the NS-trained model; prototype fusion therefore does not participate in checkpoint selection. Neurosymbolic training activates the rule after 20 warm-up epochs, initializes the prototypes from one non-shuffled pass over the training set, and then jointly optimizes all model parameters. Early stopping uses patience 10 after selection becomes active. The same rule parameters are used throughout: $\lambda_{\mathrm{rule}}=0.1$, margin $m=0.1$, temperature $\tau=0.25$, and inference-fusion coefficient $\beta=0.5$ for \FullNS{} (zero for \RegOnly{}).

The label-scarcity experiment uses a fixed budget of 150 optimization updates with 60 warm-up updates, batch size 64, evaluation batch size 512, learning rate $2\times10^{-3}$, weight decay $10^{-4}$, and gradient-norm clipping at 5. The same four architectures and the same neurosymbolic hyperparameters are retained. Model seed 1 is fixed so that the intended source of repetition in this experiment is the support-set realization rather than repeated neural initializations.

\subsection{Spatial and stratified evaluation protocols}
\label{subsec:protocols}

\begin{figure}[htbp]
\centering

\begin{minipage}[c]{0.48\textwidth}
    \centering
    \includegraphics[
        width=0.4\linewidth,
        trim=40 300 40 300,
        clip
    ]{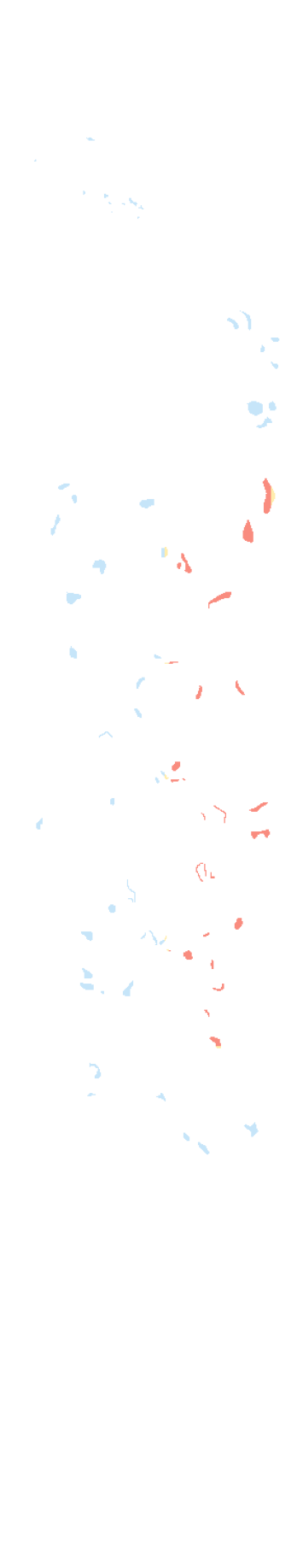}\\[-0.2em]
    \textbf{(a) Botswana (central crop)}
\end{minipage}
\hfill
\begin{minipage}[c]{0.48\textwidth}
    \centering
    \includegraphics[
        height=0.25\textheight,
        keepaspectratio
    ]{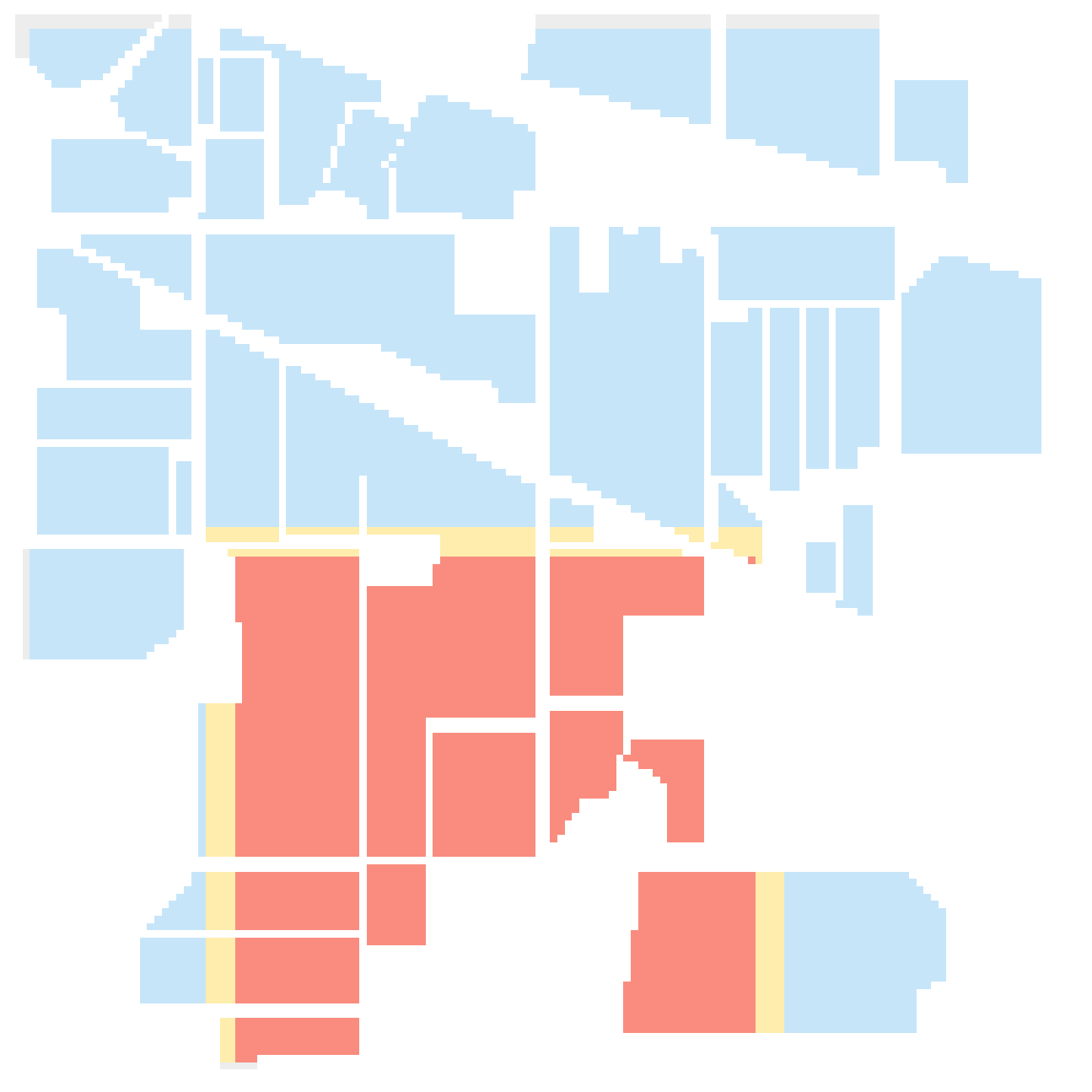}\\[-0.2em]
    \textbf{(b) Indian Pines}
\end{minipage}

\vspace{0.8em}

\begin{minipage}[c]{0.48\textwidth}
    \centering
    \includegraphics[height=0.25\textheight,keepaspectratio]
    {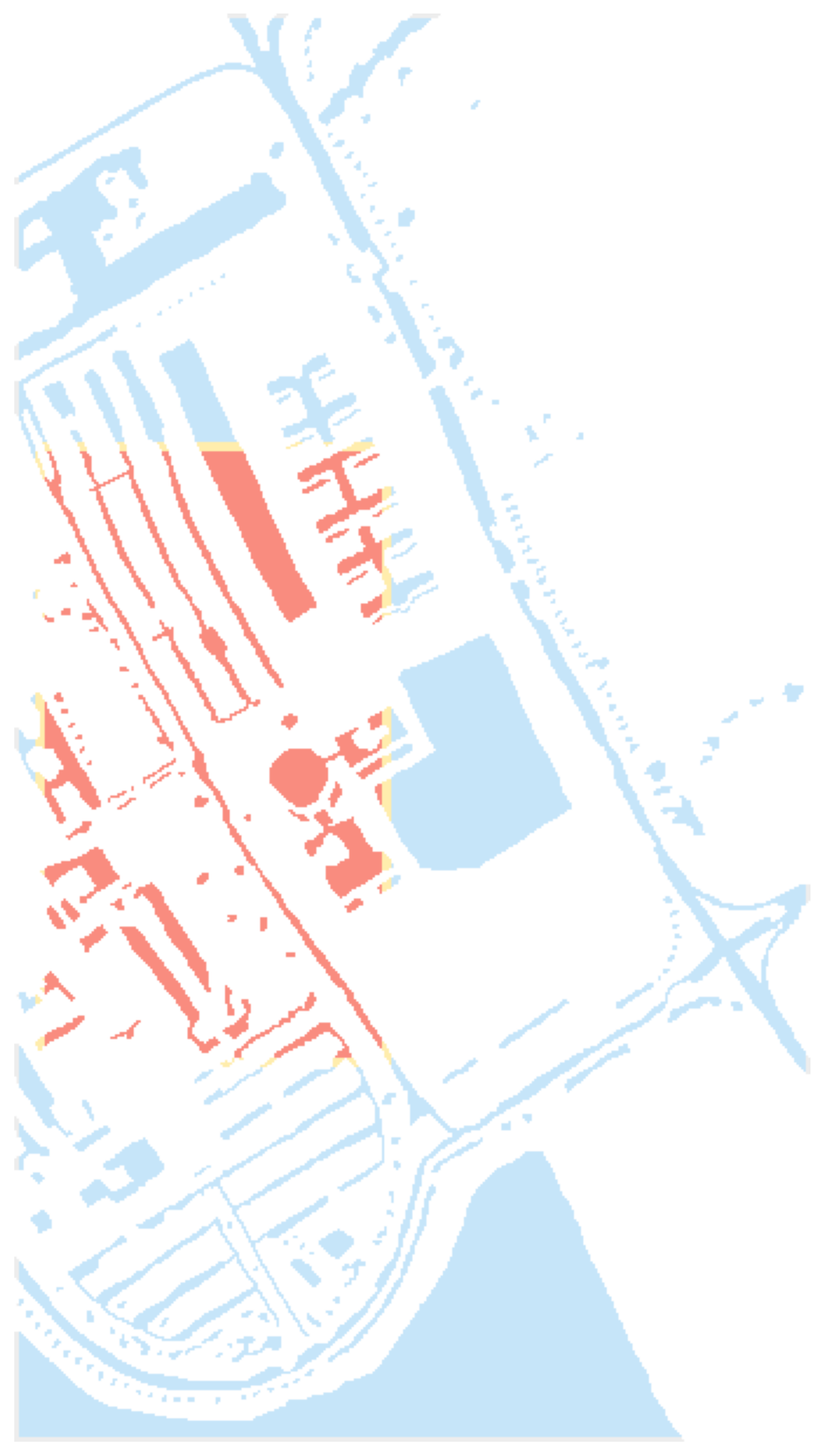}\\[-0.2em]
    \textbf{(c) Pavia University}
\end{minipage}
\hfill
\begin{minipage}[c]{0.48\textwidth}
    \centering
    \includegraphics[height=0.25\textheight,keepaspectratio]
    {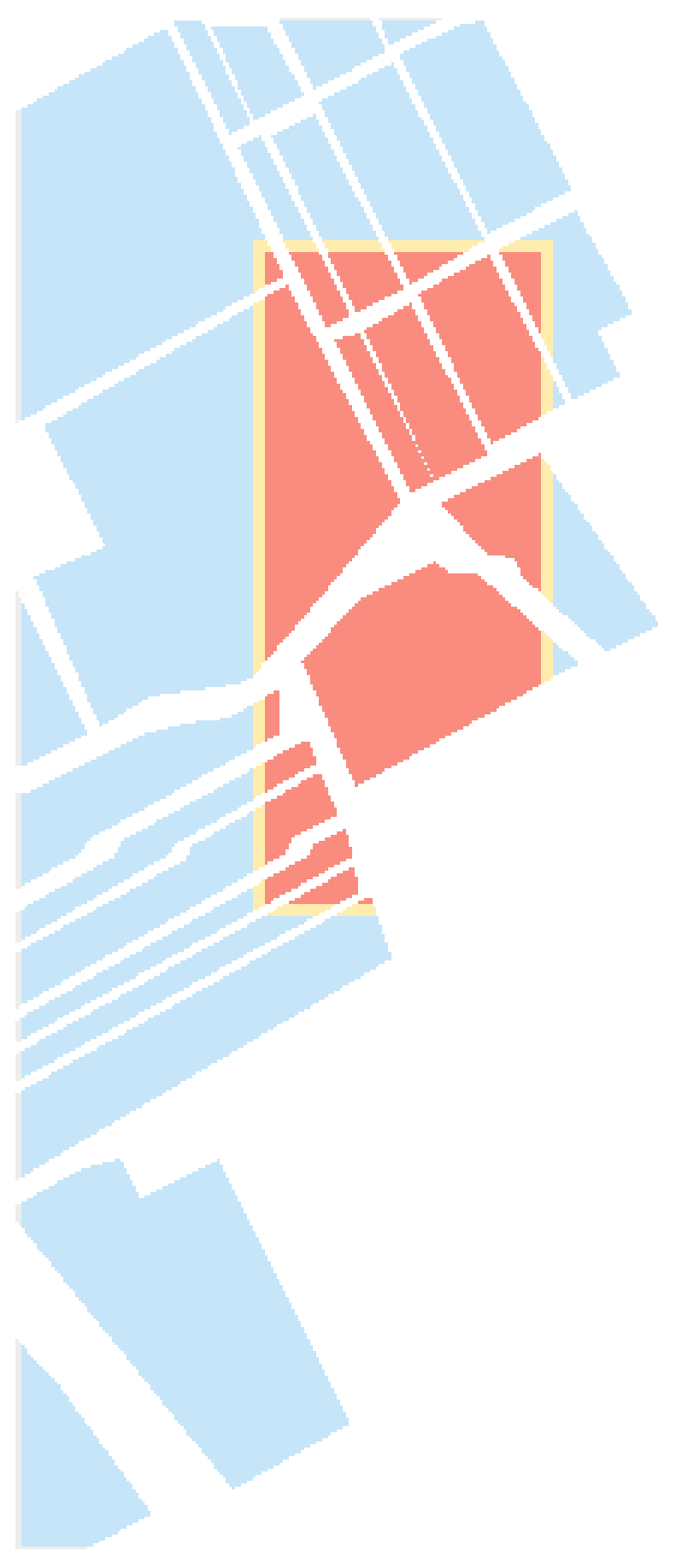}\\[-0.2em]
    \textbf{(d) Salinas}
\end{minipage}

\caption{
Spatial split configurations used for leakage-controlled evaluation:
(a) Botswana, (b) Indian Pines, (c) Pavia University, and
(d) Salinas. Blue and green indicate the selected training and validation samples, respectively, drawn from spatially separated eligible regions. Red denotes all valid labeled samples within the spatially held-out test block. Yellow indicates the spatial dead zone introduced to prevent patch overlap between split roles. Light-gray pixels correspond to labeled samples not selected for training, validation, or testing in the illustrated fold, whereas white denotes unlabeled background. Owing to the highly elongated geometry of the Botswana scene, panel (a) shows an enlarged central crop for
visual clarity; the split itself is constructed on the complete scene.
}
\label{fig:spatial_splits}
\end{figure}

The primary leakage-aware protocol constructs spatially separated train, validation, and test regions and applies a four-pixel dead-zone/buffer between split roles \cite{gkologkinas2026spatially}. Because the input is a $5\times5$ patch, this buffer prevents patch footprints from overlapping and, thus, reduces the spatial dependence across splits.
The number of feasible spatial folds is constrained by scene geometry and class coverage: the executed standard evaluation contains three folds for Pavia University and two folds each for Botswana, Indian Pines, and Salinas. Figure \ref{fig:spatial_splits} demonstrates the proposed area separation.

Botswana required a more permissive validation-block search because of its sparse labeled geometry: the target validation fraction was reduced from 0.10 to 0.06, the minimum validation class-coverage fraction from 0.45 to 0.25, and the minimum validation-patch count from 30 to 10; smaller candidate block scales (0.55 and 0.65) were also admitted. These changes affect split feasibility only and do not alter the Rank-R or neurosymbolic objectives. With so few independent spatial partitions, these runs are summarized descriptively rather than converted into underpowered fold-level significance claims.

A secondary seven-fold stratified protocol is retained for comparability with conventional Rank-R evaluation. Labeled centers are disjoint between train, validation, and test roles, but nearby patches may remain spatially correlated. The standard stratified setup uses 20 training and 10 validation samples per class where feasible; Botswana retains 20 training samples but uses six validation samples per class because of its smaller class supports. Consequently, stratified and spatial results are reported separately and are not pooled as exchangeable replications. The distinction is central to RQ3: the stratified experiment asks how the method behaves under conventional sampling, whereas the spatial experiment provides the more conservative estimate of geographic generalization within a scene \cite{feng2023leakage,cao2023disjoint}.

\subsection{Label-scarcity protocol}
\label{subsec:scarcity}

Because the support-set size increases with $K$ while the optimization budget remains fixed at 150 updates, the effective number of passes through the support set is larger at small $K$ and smaller at large $K$. The scarcity experiment therefore controls update budget rather than epoch budget.

The scarcity study is conducted only with spatial separation. Per-class support budgets are
\begin{equation}
K\in\{2,3,5,7,10,15,20\}.
\end{equation}
For each spatial fold, five support seeds (101, 202, 303, 404, and 505) generate nested support sets, so that a larger $K$ extends rather than replaces the support ordering associated with that seed. This pairing is essential because changes with $K$ are then less contaminated by unrelated resampling. The experiment contains 3,360 method-level evaluations (2,240 trained models, since RankR-NS and RankR-NS-RegOnly share a checkpoint) across datasets, spatial folds, support seeds, budgets, architectures, and the three method variants. 

The usable scarcity-fold counts are three for Pavia University, two for Botswana, two for Salinas, and one for Indian Pines. $K=1$ is intentionally excluded from the main study because a one-sample class prototype is identical to its sole support embedding and therefore ceases to represent an aggregate class-level relation.

\subsection{Evaluation and statistical analysis}
\label{subsec:stats}

\MacroF{} is the primary response because the datasets are class imbalanced and the neurosymbolic rule is defined at class level. For each evaluation fold, Macro-F1 is calculated as
the unweighted mean of the class-specific F1 scores, considering only classes with non-zero ground-truth support in that fold. Overall accuracy is retained as a secondary descriptive metric in the archived results but is not used to drive the paper's inferential claims. All method comparisons are paired within the same split and architecture.

For the seven-fold stratified protocol, the dataset-level analysis first averages architecture-specific scores within each matched fold. A Friedman test compares the three methods; when the omnibus test is significant, paired Wilcoxon signed-rank post-hoc comparisons are performed with Holm correction \cite{friedman1937ranks,wilcoxon1945individual,holm1979simple,demsar2006statistical}. Paired direction is summarized with mean and median $\Delta$\MacroF{}, win/tie/loss counts, and rank-biserial effect size. Architecture-specific analyses use the same gatekeeping logic. For the spatial protocol, the two or three available folds per dataset are insufficient for informative rank-based tests, so only paired magnitudes and directions are interpreted. These descriptive quantities are retained in the archived paired-comparison outputs and accompany the Wilcoxon/Holm results used for the stratified analysis.

For label scarcity, the primary descriptive quantity is the paired difference
\begin{equation}
\Delta\mathrm{F1}(K)=\mathrm{MacroF1}_{\mathrm{NS}}(K)-\mathrm{MacroF1}_{\mathrm{RankR}}(K).
\label{eq:gain}
\end{equation}
Cross-dataset summaries first average within each dataset and architecture and then weight the four datasets equally, preventing the much larger Salinas and Pavia scenes from dominating the summary. To test whether the gain changes systematically with label availability without treating support seeds as independent spatial replications, a sensitivity analysis first averages support-seed repetitions within each independent spatial fold and architecture and then estimates the slope of paired gain against $\log_2 K$. Thus, a slope is interpretable as the change in gain for each doubling of labeled samples per class. Architecture-wise fold slopes are tested against zero with Wilcoxon signed-rank tests and Holm correction.

A mixed-effects formulation was also explored in the archived analysis. For the two primary contrasts against RankR, however, the fitted random-effect covariance collapsed to a singular boundary and the Hessian was not positive definite, producing unstable standard errors and confidence intervals. Those model-based intervals are therefore not used as evidence in this paper. The reported scarcity conclusions rely on the observed paired gains and the fold-level slope sensitivity analysis rather than on a numerically invalid fit.

\section{Results}
\label{sec:results}
The results are organized to test whether the prototype-rule effect is stable across the conditions introduced in Section \ref{sec:methodology}, or whether it depends on how and where it is measured. Section \ref{subsec:results-standard} reports dataset-level changes under the spatial and stratified protocols, which already show considerable heterogeneity; Section \ref{subsec:architecture-results} shows that this heterogeneity persists, and in places reverses, at the level of individual architectures. Section \ref{subsec:scarcity_impact} turns to the label-scarcity study, asking whether the effect strengthens as supervision decreases.

\subsection{Standard evaluation: dataset-level effects}
\label{subsec:results-standard}

\begin{figure}[t]
\centering
\includegraphics[width=0.94\linewidth]{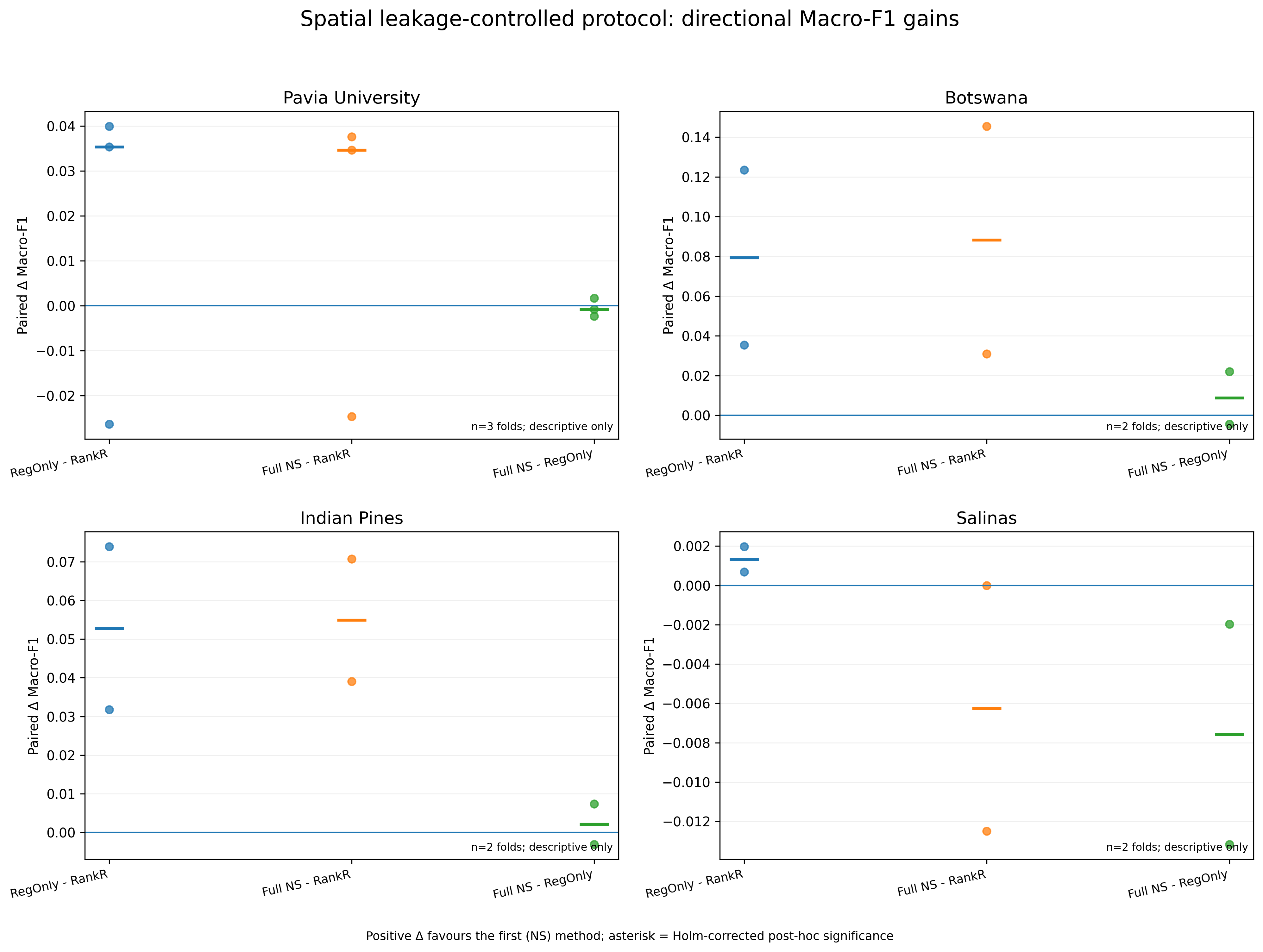}
\caption{Paired dataset-level \MacroF{} changes under the spatially separated protocol. The small number of spatial folds is treated descriptively; the figure emphasizes effect direction and magnitude rather than fold-level significance.} 
\label{fig:directional-spatial}
\end{figure}

Table ~\ref{tab:dataset-deltas} reports architecture-averaged paired changes in \MacroF{} for the three mechanistic contrasts. Under spatial evaluation, full neurosymbolic inference improves on RankR in three of the four datasets: +8.82 percentage points (pp) on Botswana, +5.49 pp on Indian Pines, and +1.59 pp on Pavia University. Salinas is the exception, with a -0.62 pp change. The regularization-only contrast is positive in all four spatial datasets, ranging from +0.13 pp on Salinas to +7.94 pp on Botswana. The difference between \FullNS{} and \RegOnly{} is comparatively small: +0.88 pp on Botswana, +0.21 pp on Indian Pines, -0.05 pp on Pavia University, and -0.76 pp on Salinas. These values answer RQ2 more directly than the end-to-end contrast alone: most of the spatial improvement is already present before prototype fusion is applied. The direction and magnitude of these spatial contrasts are summarized visually in Fig. ~\ref{fig:directional-spatial}.

\begin{table}[t]
\centering
\caption{Absolute dataset-level \MacroF{} (mean $\pm$ SD, percentage points). For each method, architecture-specific scores are first averaged within each matched fold and the resulting fold-level values are then summarized across folds.}
\label{tab:absolute-performance}
\small
\resizebox{\linewidth}{!}{%
\begin{tabular}{llccc}
\toprule
Protocol & Dataset & RankR & \RegOnly{} & \FullNS{} \\
\midrule
\multirow{4}{*}{Spatial}
 & Pavia University & $62.57\pm15.28$ & $64.20\pm18.15$ & $64.16\pm17.96$ \\
 & Indian Pines     & $24.91\pm2.26$  & $30.19\pm0.72$  & $30.40\pm0.02$ \\
 & Salinas          & $58.04\pm9.24$  & $58.18\pm9.15$  & $57.42\pm8.36$ \\
 & Botswana         & $48.64\pm5.21$  & $56.58\pm1.02$  & $57.46\pm2.88$ \\
\midrule
\multirow{4}{*}{Stratified}
 & Pavia University & $68.04\pm1.36$  & $69.41\pm1.50$  & $69.23\pm1.61$ \\
 & Indian Pines     & $51.77\pm1.68$  & $52.31\pm1.08$  & $51.82\pm0.90$ \\
 & Salinas          & $81.77\pm1.15$  & $80.64\pm2.00$  & $80.35\pm1.97$ \\
 & Botswana         & $84.36\pm1.83$  & $85.83\pm0.57$  & $85.89\pm0.72$ \\
\bottomrule
\end{tabular}}
\end{table}
Table ~\ref{tab:absolute-performance} provides the corresponding absolute performance levels and fold-to-fold dispersion, complementing the paired changes reported in Table ~\ref{tab:dataset-deltas}.
\begin{table}[t]
\centering
\caption{Dataset-level paired change in \MacroF{} (percentage points). Architecture scores are averaged within matched folds before contrasts are formed. Positive values favor the method named first. An asterisk (*) denotes a Holm-corrected post-hoc $p<0.05$ in the seven-fold stratified analysis. No inferential tests are reported for the two/three-fold spatial protocol.}
\label{tab:dataset-deltas}
\small
\resizebox{\linewidth}{!}{%
\begin{tabular}{llrrr}
\toprule
Protocol & Dataset & \RegOnly$-$RankR & \FullNS$-$RankR & \FullNS$-$\RegOnly \\
\midrule
\multirow{4}{*}{Spatial}
 & Pavia University & +1.63 & +1.59 & -0.05 \\
 & Indian Pines     & +5.28 & +5.49 & +0.21 \\
 & Salinas          & +0.13 & -0.62 & -0.76 \\
 & Botswana         & +7.94 & +8.82 & +0.88 \\
\midrule
\multirow{4}{*}{Stratified}
 & Pavia University & +1.36 & +1.18 & -0.18$^{*}$ \\
 & Indian Pines     & +0.54 & +0.05 & -0.49 \\
 & Salinas          & -1.13 & -1.42$^{*}$ & -0.29 \\
 & Botswana         & +1.46 & +1.52 & +0.06 \\
\bottomrule
\end{tabular}}
\end{table}

\begin{figure}[t]
\centering
\includegraphics[width=0.94\linewidth]{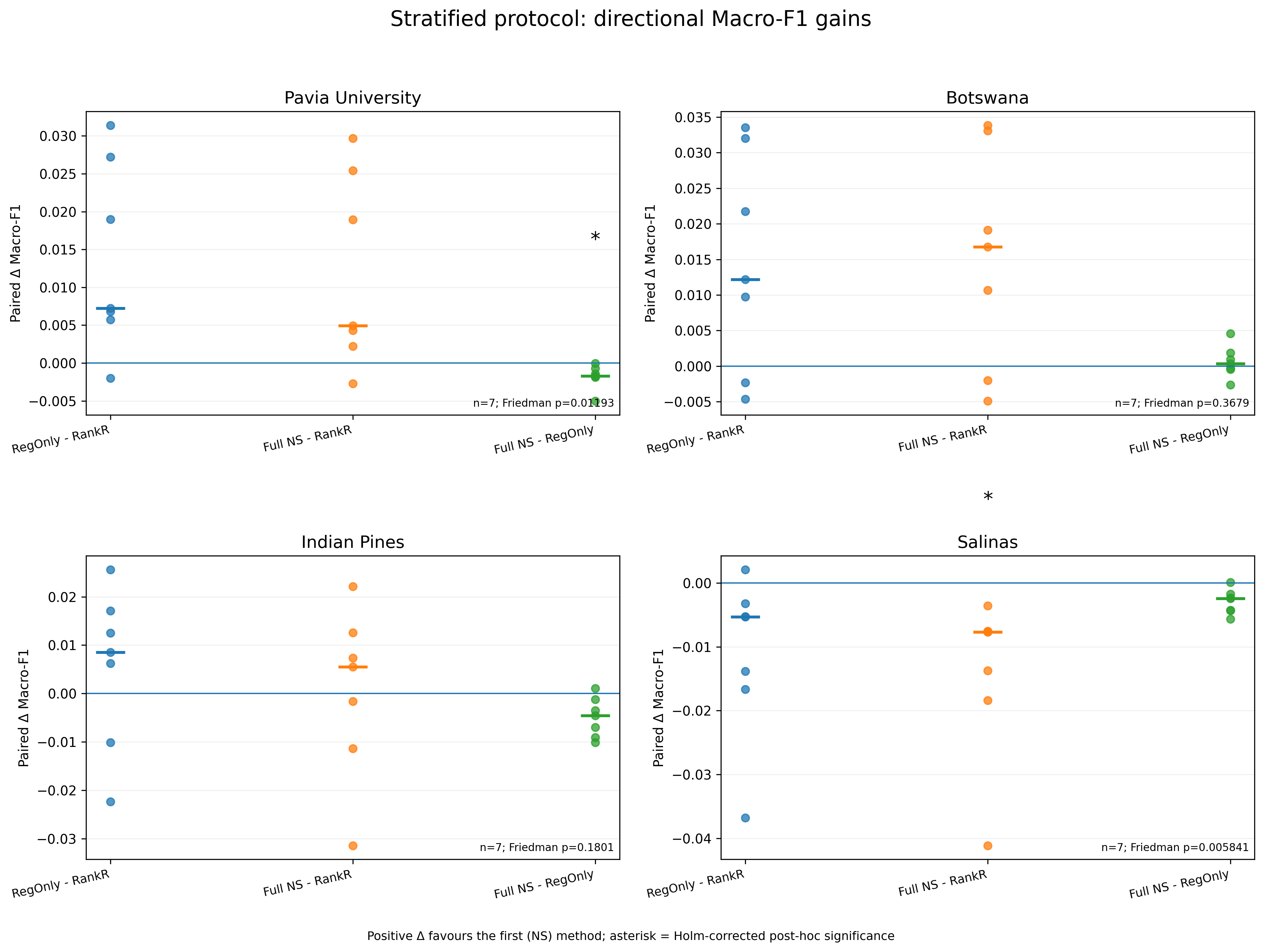}
\caption{Paired dataset-level \MacroF{} changes under seven-fold stratified evaluation. Post-hoc testing is performed only when the dataset-level Friedman omnibus test is significant.} 
\label{fig:directional-stratified}
\end{figure}

The conventional stratified protocol exhibits the same broad heterogeneity but smaller end-to-end effects. \FullNS{} is positive on Pavia University (+1.18 pp) and Botswana (+1.52 pp), essentially neutral on Indian Pines (+0.05 pp), and negative on Salinas (-1.42 pp). The Pavia University Friedman test is significant ($p=0.0119$), but the baseline-to-NS post-hoc contrasts do not survive Holm correction ($p_{\mathrm{Holm}}=0.0625$ for both \RegOnly{} and \FullNS{}). The fusion contrast is negative in all seven Pavia folds and does survive correction ($p_{\mathrm{Holm}}=0.0469$), showing that the prototype fusion step slightly erodes the gain produced during training. Salinas also has a significant omnibus test ($p=0.00584$); \FullNS{} is below RankR in all seven folds and the post-hoc contrast survives Holm correction ($p_{\mathrm{Holm}}=0.0469$). Botswana and Indian Pines do not pass the dataset-level Friedman gate, so no post-hoc significance claim is made for them. Figure ~\ref{fig:directional-stratified} provides the corresponding dataset-level stratified contrasts.

\subsection{Architecture dependence}
\label{subsec:architecture-results}

\begin{figure}[t]
\centering
\begin{minipage}{0.49\linewidth}
\centering
\includegraphics[width=\linewidth]{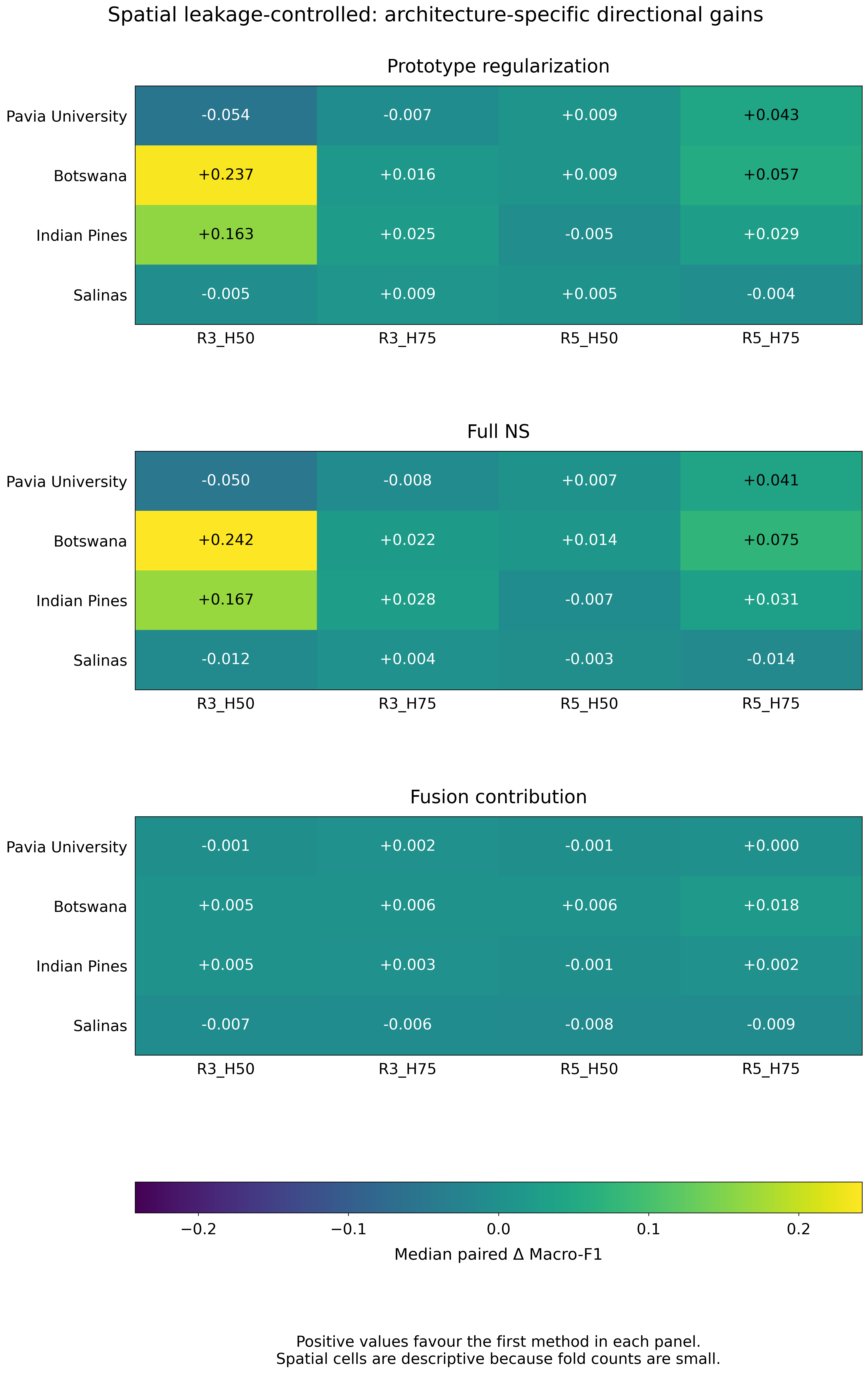}
\end{minipage}\hfill
\begin{minipage}{0.49\linewidth}
\centering
\includegraphics[width=\linewidth]{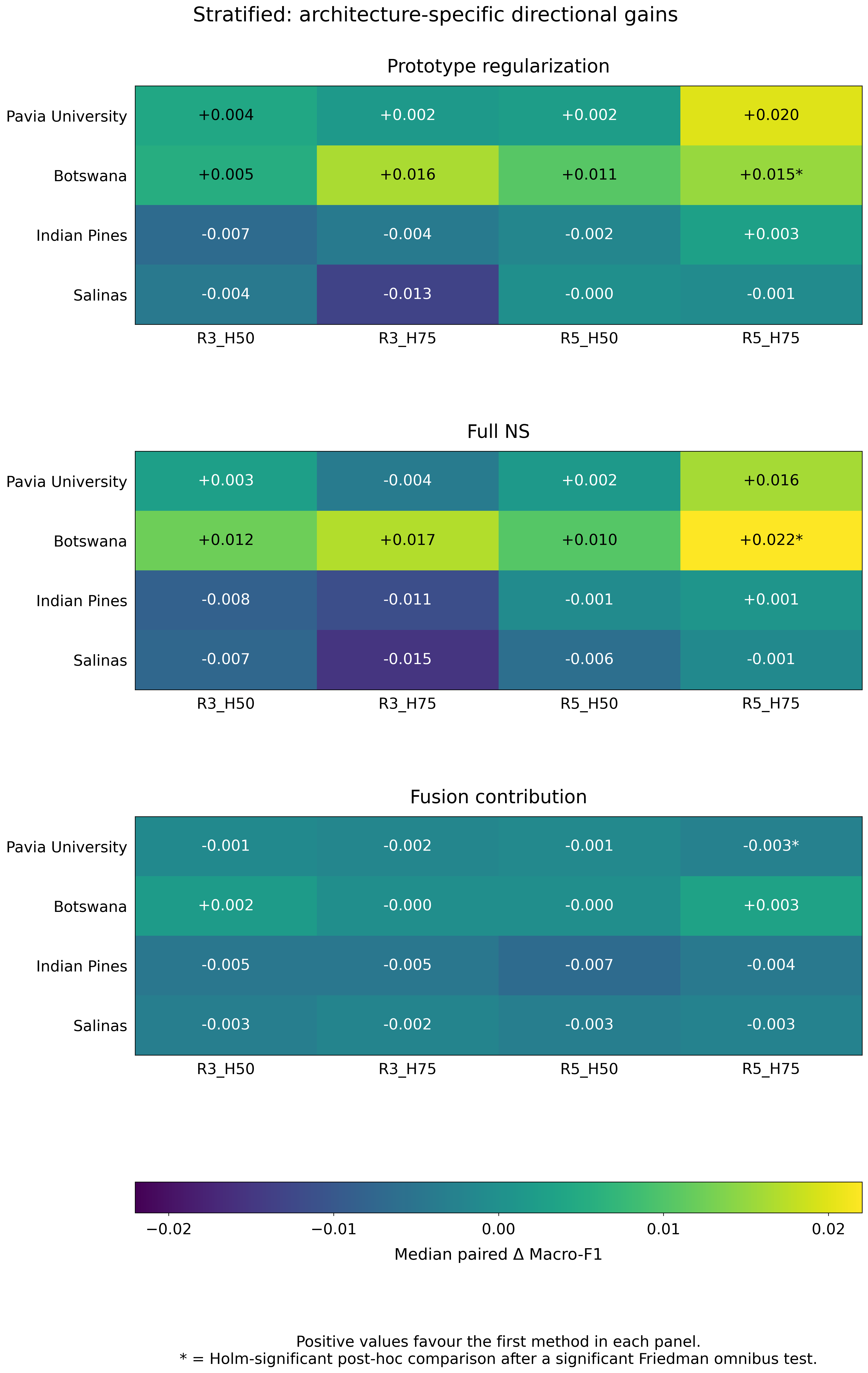}
\end{minipage}
\caption{Architecture-level directional \MacroF{} effects for spatial (left) and stratified (right) evaluation. The heatmaps show that architecture averaging can conceal changes in both magnitude and direction.}
\label{fig:architecture-heatmaps}
\end{figure}

The dataset averages conceal substantial architecture dependence (Fig.  ~\ref{fig:architecture-heatmaps}). Pavia University is the clearest example under spatial evaluation: the regularization effect is negative for $R=3$ (mean changes of -3.56 pp at $H=50$ and -1.71 pp at $H=75$) but positive for $R=5$ (+3.60 and +8.20 pp, respectively). This reversal explains why the dataset - level mean is only moderately positive despite large gains in some configurations. Botswana is also heterogeneous in magnitude, with particularly large spatial gains in R3-H50, but its dataset - level direction remains positive.

The stratified architecture analysis likewise contains effects that are diluted by architecture averaging. For Botswana with $R=5,H=75$, both \RegOnly{} and \FullNS{} improve over RankR in all seven folds; the mean gains are +2.73 and +3.02 pp, and both paired comparisons survive Holm correction ($p_{\mathrm{Holm}}=0.0469$). Conversely, Pavia University at $R=5,H=75$ shows a significant negative fusion contribution with the same corrected $p$-value. These results answer RQ5: the prototype-rule effect is not solely dataset-specific; it interacts with the capacity/structure of the Rank-R representation.

\subsection{Scarcity impact}
\label{subsec:scarcity_impact}

\begin{figure}[t]
\centering
\includegraphics[width=0.88\linewidth]{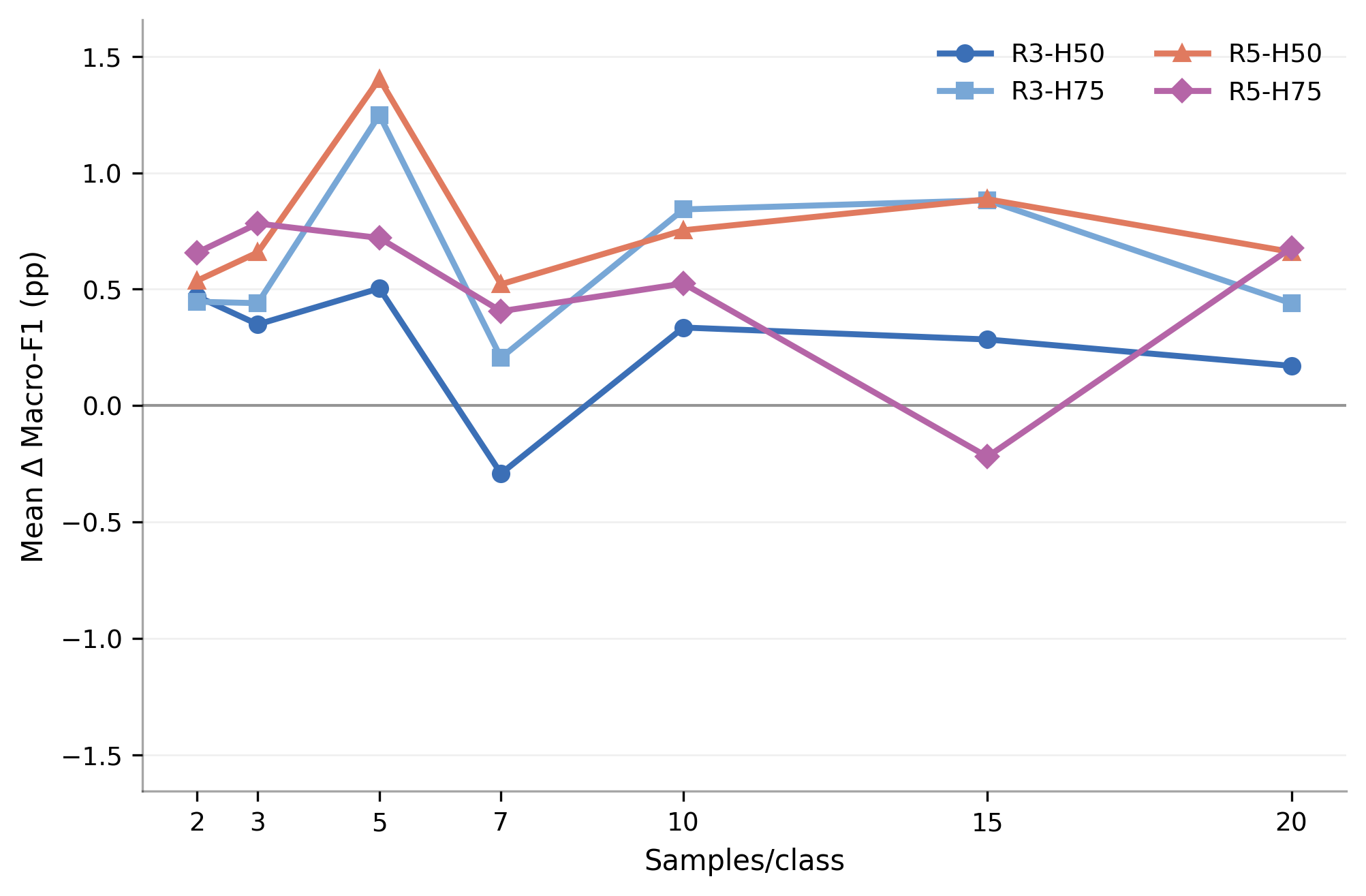}
\caption{Equal-dataset-weighted full-neurosymbolic \MacroF{} gain over RankR across label budgets and Rank-R architectures. The overall gain is usually positive but not monotonic in $K$.}
\label{fig:scarcity-fullns}
\end{figure}

The scarcity experiment does not support a simple monotonic statement that neurosymbolic regularization becomes progressively more useful as labels disappear. After equal weighting of the four datasets and averaging across the four architectures, the \FullNS{} gain over RankR is positive at every tested budget but varies non-monotonically: +0.53 pp at $K=2$, +0.56 at $K=3$, +0.97 at $K=5$, +0.21 at $K=7$, +0.61 at $K=10$, +0.46 at $K=15$, and +0.49 at $K=20$ (Table ~\ref{tab:scarcity-summary}). \RegOnly{} is also positive on average at all budgets, with smaller gains of +0.24--+0.62 pp. Taking the difference between the two rows of Table \ref{tab:scarcity-summary}, the average incremental contribution of fusion is modest, ranging from $-0.03$ pp at $K=7$ to $+0.35$ pp at $K=5$. The equal-dataset-weighted full-neurosymbolic trajectory is shown in Fig. ~\ref{fig:scarcity-fullns}.

\begin{table}[t]
\centering
\caption{Equal-dataset-weighted paired \MacroF{} gain over RankR across the label budgets. Values are percentage points and are averaged over the four Rank-R architectures after dataset-level aggregation.}
\label{tab:scarcity-summary}
\small
\begin{tabular}{rrrrrrrr}
\toprule
$K$ & 2 & 3 & 5 & 7 & 10 & 15 & 20 \\
\midrule
\FullNS$-$RankR & 0.53 & 0.56 & 0.97 & 0.21 & 0.61 & 0.46 & 0.49 \\
\RegOnly$-$RankR & 0.31 & 0.43 & 0.62 & 0.24 & 0.38 & 0.39 & 0.42 \\
\bottomrule
\end{tabular}
\end{table}

Dataset-level behavior explains the irregular aggregate curve. Botswana consistently benefits from \FullNS{}, with architecture-averaged gains between +1.10 and +2.92 pp across all seven budgets. Pavia University is positive from $K=2$ through $K=10$ (+0.21 to +0.78 pp), is approximately neutral at $K=20$ (+0.03 pp), and is slightly negative at $K=15$ (-0.20 pp). Indian Pines is mixed, ranging from -0.27 to +1.02 pp depending on $K$. Salinas is negative at six of seven budgets; its only positive architecture-averaged value is at $K=5$ (+0.51 pp). The same dataset dependence was already visible in the standard evaluation, indicating that label budget alone is not the dominant moderator.

The fold-level slope sensitivity analysis provides a more direct test of RQ4. For \FullNS{} versus RankR, 22 of 32 architecture-by-spatial-fold slopes are negative and 10 are positive when gain is regressed on $\log_2 K$. A negative slope means the NS gain shrinks as labels increase, i.e., the gain is larger when labels are scarcer.

The mean slopes are -0.137, -0.039, -0.009, and -0.156 pp per doubling of $K$ for R3-H50, R3-H75, R5-H50, and R5-H75, respectively. R5-H75 has seven negative slopes out of eight, but its Holm-corrected Wilcoxon $p$-value is $p_{\mathrm{Holm}}=0.2188$; the other three architecture-wise tests are also non-significant after correction. Thus the directional imbalance is suggestive, but the experiments do not establish a general label-budget trend at the conventional 0.05 level.

\begin{figure}[t]
\centering
\begin{minipage}{0.49\linewidth}
\centering
\includegraphics[width=\linewidth]{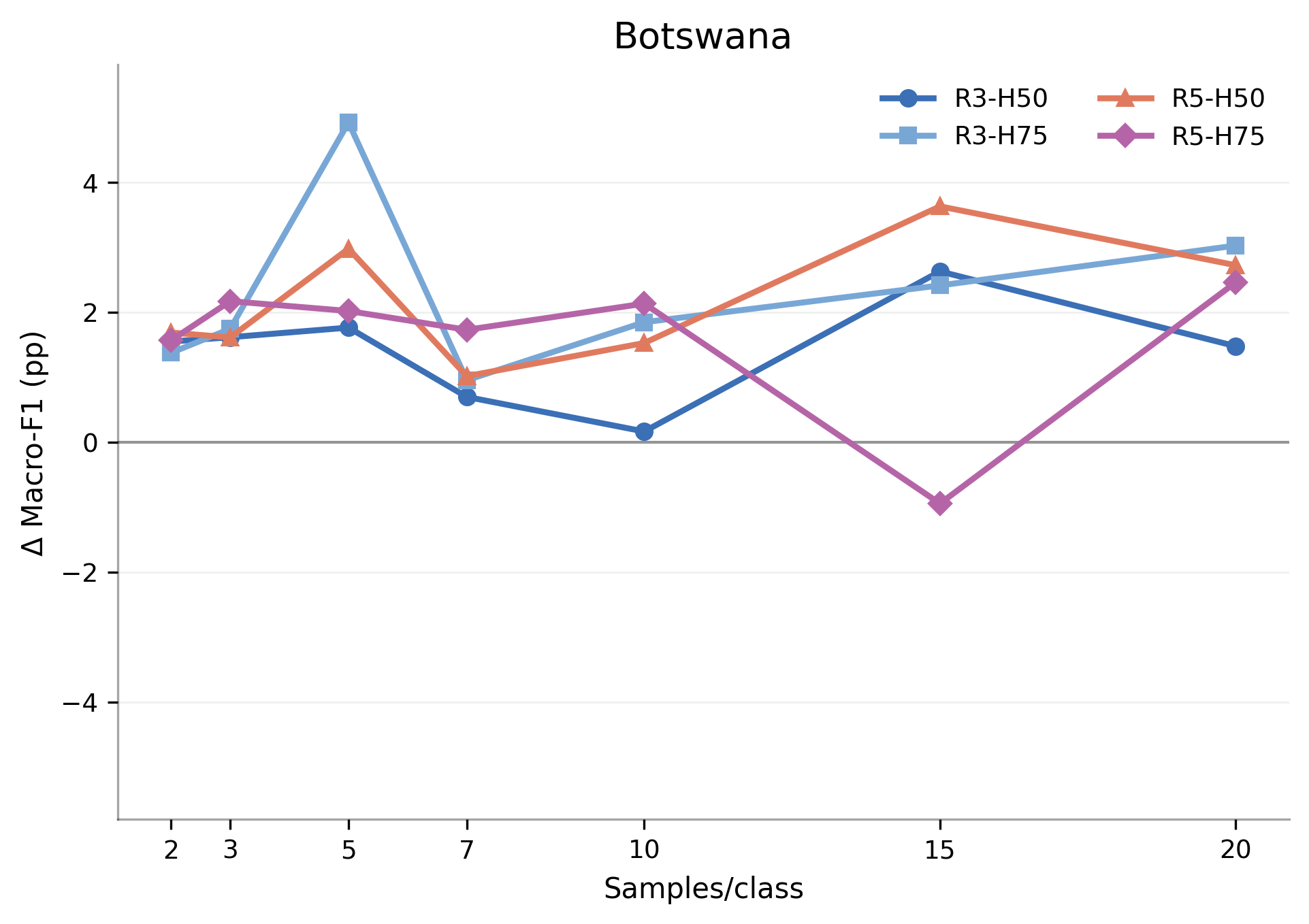}
\end{minipage}\hfill
\begin{minipage}{0.49\linewidth}
\centering
\includegraphics[width=\linewidth]{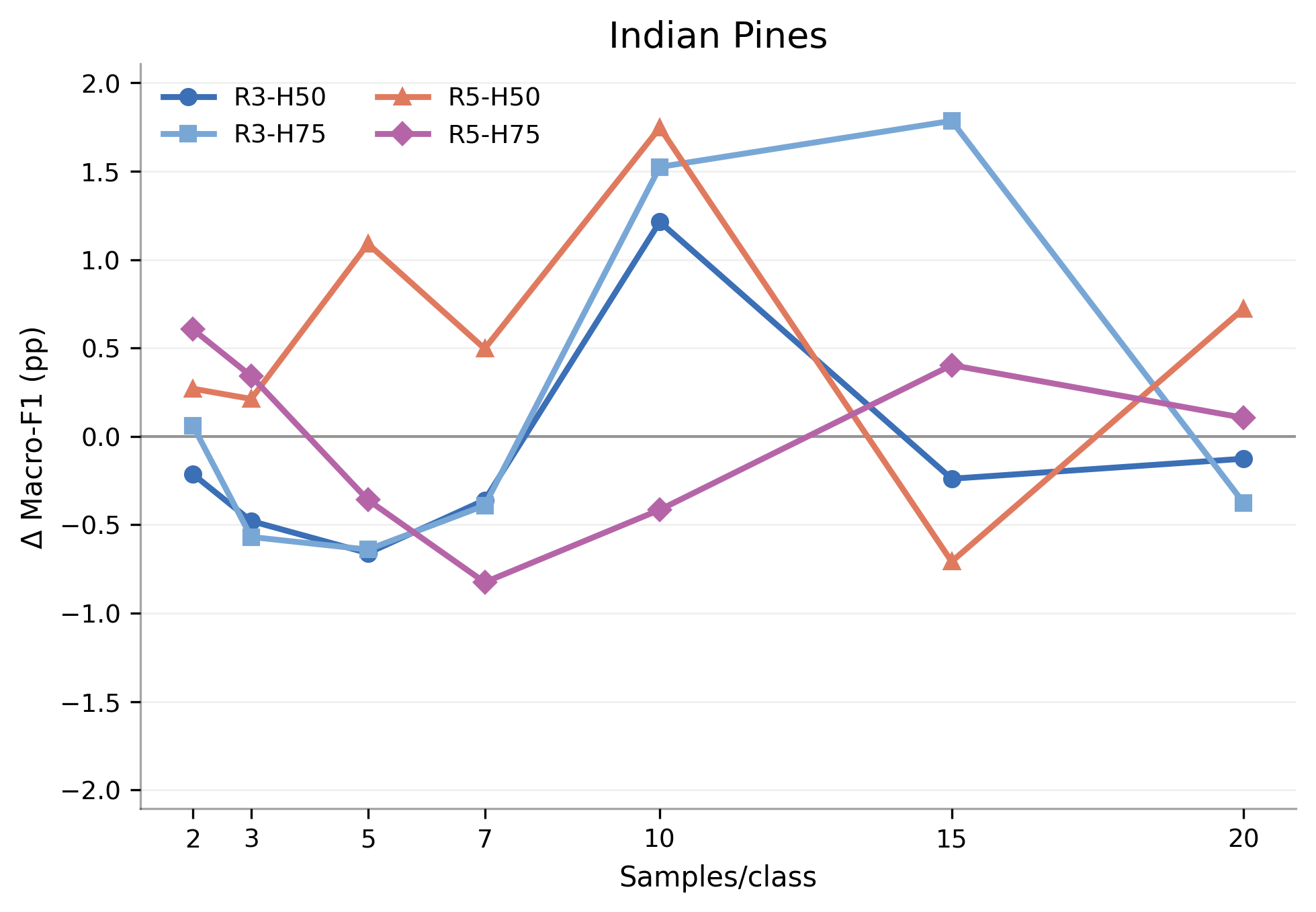}
\end{minipage}
\\[0.5em]
\begin{minipage}{0.49\linewidth}
\centering
\includegraphics[width=\linewidth]{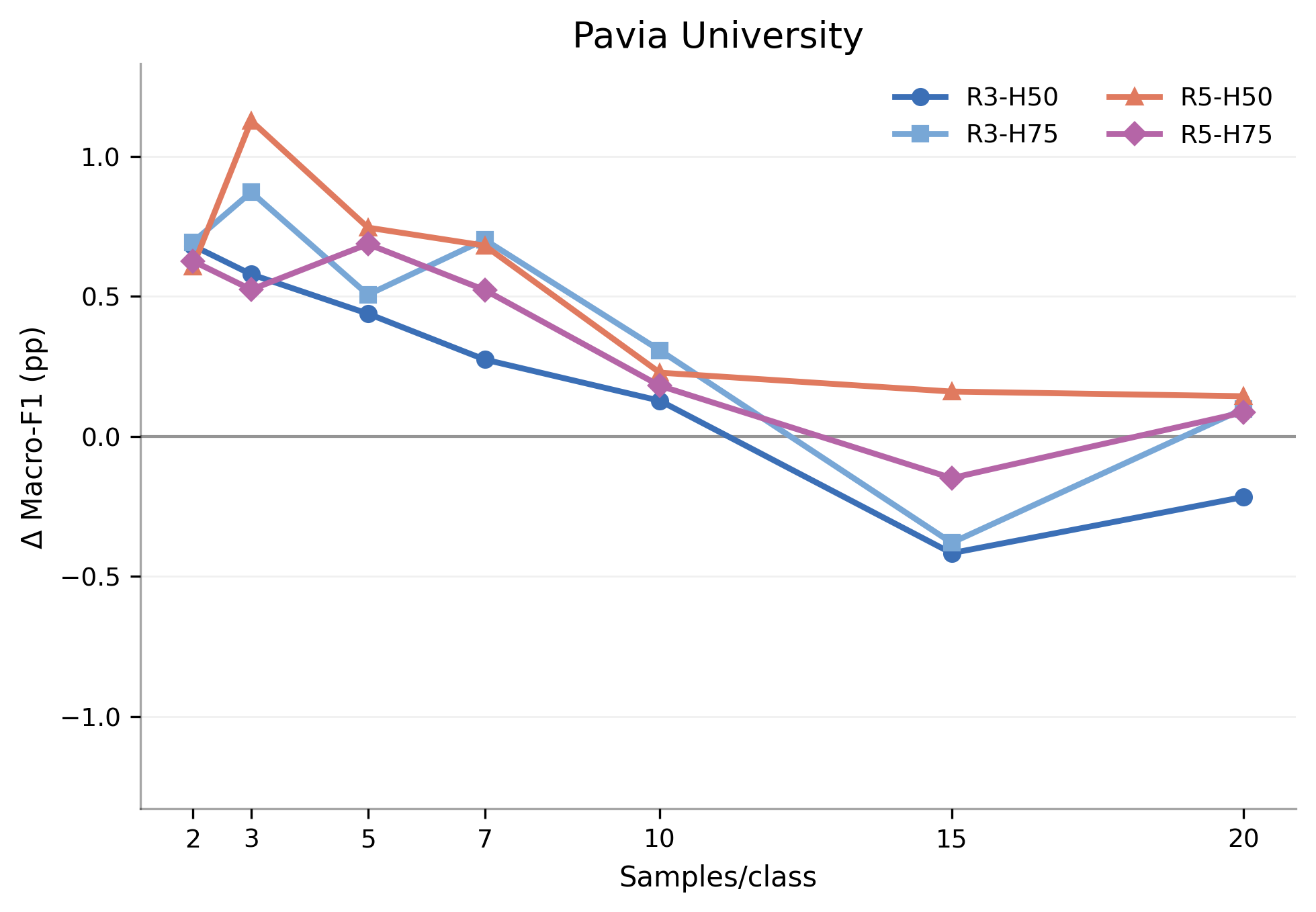}
\end{minipage}\hfill
\begin{minipage}{0.49\linewidth}
\centering
\includegraphics[width=\linewidth]{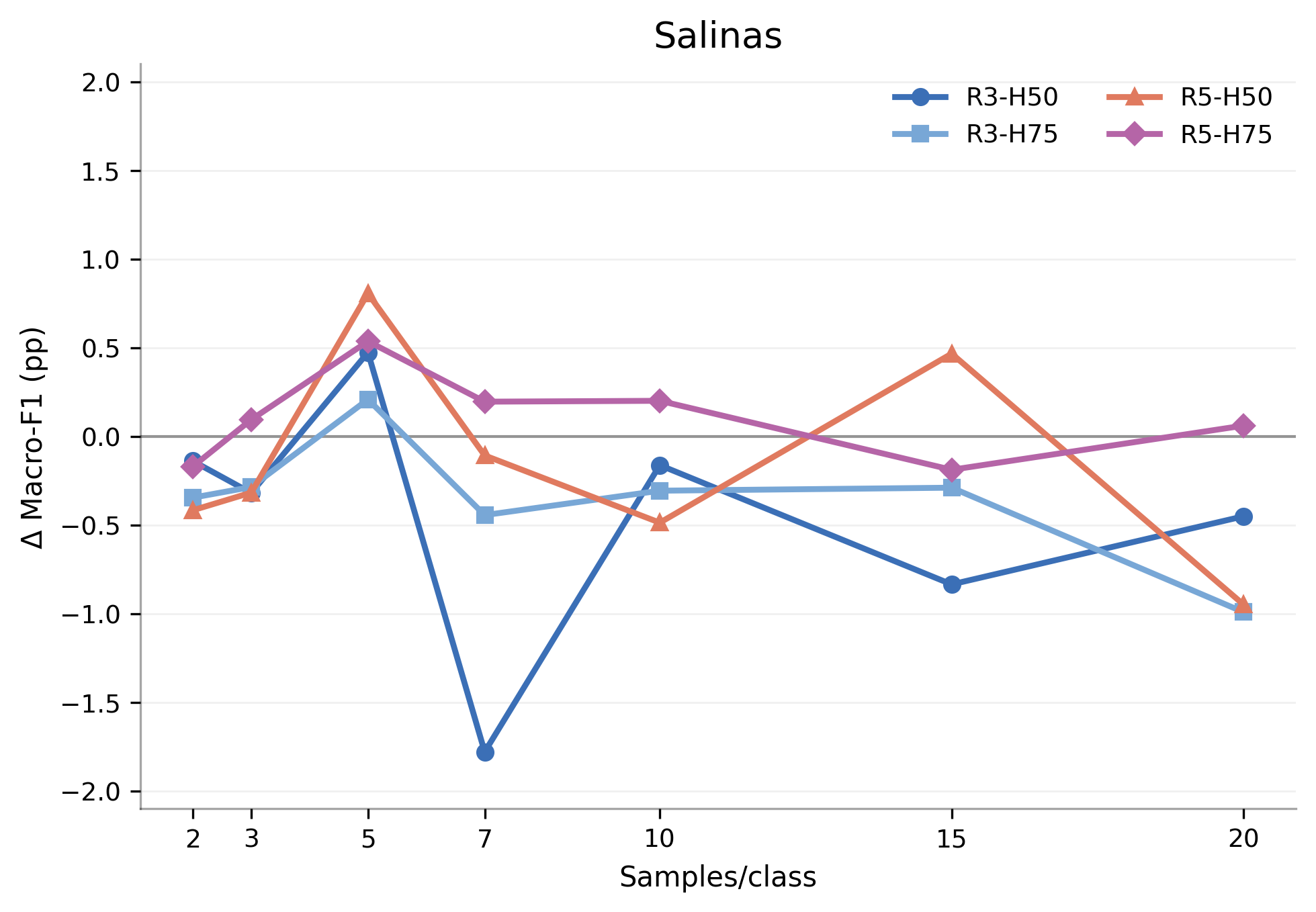}
\end{minipage}
\caption{Dataset-specific \FullNS{} gain over RankR across label budgets. The principal result is heterogeneity rather than a universal scarcity curve.}
\label{fig:scarcity-datasets}
\end{figure}

Figure ~\ref{fig:scarcity-datasets} makes the heterogeneity explicit: the same architecture can change direction across datasets and budgets, and no single low-$K$ regime dominates uniformly.

\section{Discussion}
\label{sec:discussion}
In this section we discuss the results in relation to the five research questions. Section \ref{subsec:discussion-when} addresses RQ1 and RQ3, arguing that the neurosymbolic effect is conditional rather than universal and depends on the evaluation protocol. Section \ref{subsec:discussion-bias} addresses RQ2, attributing most of the effect to training-time regularization rather than inference-time fusion, while Section \ref{subsec:discussion-scarcity} addresses RQ4 and RQ5, showing that the effect of label scarcity is not monotonic and interacts with dataset and architecture. Section \ref{subsec:discussion-symbolic} clarifies the intended, restricted sense in which the mechanism is called neurosymbolic, and Section \ref{subsec:limitations} states the study's limitations.

\subsection{A conditional rather than universal neurosymbolic effect}
\label{subsec:discussion-when}

The experiments answer RQ1 with a qualified result. Prototype-rule regularization can materially improve a Rank-R model, but the sign and magnitude of the effect depend on the dataset and architecture. This is most visible in the spatial evaluation, where the end-to-end \FullNS{} effect ranges from +8.82 pp on Botswana to -0.62 pp on Salinas. The result is therefore stronger than a claim of negligible average change but weaker than a claim of universal benefit. Such heterogeneity is also visible in the stratified protocol, where Salinas is the only dataset for which full NS is consistently below RankR in all seven folds and the post-hoc comparison survives Holm correction.

The fact that protocol changes can alter the magnitude and, for some contrasts, the apparent direction is also important for RQ3. Stratified folds provide more replications and support non-parametric testing, but they do not reproduce the geographic independence of the spatial folds. The spatial experiment should therefore be treated as the primary evidence about leakage-controlled within-scene generalization, while the stratified experiment serves as a conventional comparison. A method whose advantage is visible only under one sampling scheme would warrant caution; here, the broad dataset pattern (strong Botswana, positive Indian/Pavia in several settings, weak or negative Salinas) persists sufficiently to indicate genuine dataset dependence rather than a single protocol artifact.

\subsection{Training-time regularization is the main mechanism}
\label{subsec:discussion-bias}

RQ2 is addressed by the shared-checkpoint ablation. Across the spatial datasets, \RegOnly{} already accounts for most of the difference from RankR. Full inference fusion adds less than one percentage point in absolute \MacroF{} on every dataset and is negative on Pavia University and Salinas. Under stratified evaluation, fusion is negative in all seven Pavia folds and significantly reduces the regularization-only result after Holm correction. In the label-scarcity study, the equal-weighted difference between \FullNS{} and \RegOnly{} is likewise small relative to the full contrast with RankR.

This pattern suggests that the primary value of the prototype rule is not a second classifier layered on top of the network. Instead, it acts mainly as a training constraint on the geometry of the latent representation. This interpretation is consistent with the intended complementarity of the two inductive biases: the Rank-R decomposition restricts the tensor mapping, while the prototype rule constrains class organization in the embedding space. The inference fusion can still help in particular regimes, but the results do not justify treating it as uniformly beneficial.

\subsection{Label scarcity does not produce a simple monotonic response}
\label{subsec:discussion-scarcity}

RQ4 yields a more nuanced result than the initial hypothesis that symbolic structure should become increasingly valuable as supervision decreases. The cross-dataset gain is positive at every tested $K$, and 22 of 32 fold-level slopes point toward a larger advantage at lower budgets. Nevertheless, the aggregate curve is non-monotonic and none of the architecture-specific slope tests survives Holm correction. The strongest directional pattern occurs for R5-H75 (seven of eight negative slopes), but even there the corrected evidence is insufficient for a general claim.

The more defensible interpretation is therefore that scarcity can expose useful prototype regularization in some configurations, but it does not act as a single control variable that determines the effect. Dataset geometry, class compactness, architecture, and the quality of prototypes estimated from very small supports all plausibly interact with $K$. Salinas is instructive: reducing the label budget does not reliably turn a negative standard-evaluation effect into a positive one. Conversely, Botswana remains favorable across essentially the entire scarcity range. These observations make RQ5 at least as important as RQ4.

\subsection{Why call the mechanism neurosymbolic?}
\label{subsec:discussion-symbolic}

The prototype itself is a learned numerical parameter and is not symbolic merely because it is initialized from an average embedding. The neurosymbolic element is the explicit relation imposed on those grounded class symbols: an embedding labeled as class $c$ should satisfy a compatibility relation with the prototype representing $c$ and should be separated from prototypes representing alternative classes. That relation is named, inspectable, and added to the neural objective as a differentiable constraint. The relation is only as meaningful as the grounding of the symbols it refers to \cite{pmlr-v202-marconato23a}. Here, that grounding is minimal, since each class symbol is anchored solely to the mean of the labelled embeddings of that class and is thereafter free to drift with the representation. Because the anchoring comes from class labels rather than from unsupervised concept discovery, the rule can be satisfied only through class-consistent geometry. This is a deliberately modest use of the term neurosymbolic: the framework does not claim discrete reasoning or external domain knowledge, but it does separate a declarative class relation from the unconstrained neural classification objective.

\subsection{Limitations and implications}
\label{subsec:limitations}

Several limitations bound the conclusions. First, the standard experiments primarily use one model seed, so the independent spatial partitions provide stronger evidence about sampling variation than about variation due to neural initialization. Second, only two or three spatial folds are feasible for most datasets, and the scarcity study has only one independent Indian Pines spatial fold; support seeds improve characterization of support selection but are not substitutes for independent spatial replications. Third, a single prototype per class favors compact class geometry and may be inappropriate for multimodal classes. 
Recent few-shot HSI work has similarly questioned prototypes derived solely from a small number of support samples and explored trainable semantic anchors to enrich prototype representations \cite{yan2026}.
Fourth, the neurosymbolic hyperparameters are held fixed across datasets rather than tuned separately, which is useful for controlled comparison but may understate achievable performance in some scenes. Fifth, the knowledge grounding is induced from labeled embeddings rather than supplied by an external ontology or expert knowledge base. Finally, the study covers four hyperspectral datasets and four Rank-R architectures; it characterizes this mechanism rather than neurosymbolic learning in general.

The failed mixed-effects diagnostics also motivate a methodological point. Repeated support-set realizations create a rich empirical surface, but treating them as if they supplied many independent spatial units can produce unstable hierarchical fits. The fold-level sensitivity analysis used here is more conservative: it sacrifices nominal sample size in exchange for an inferential unit that better matches the spatial generalization question.

\section{Conclusion}
\label{sec:conclusion}

This work evaluates a simple differentiable prototype rule as a second inductive bias for Rank-R tensor neural networks. The evidence does not support a universal neurosymbolic advantage. Under spatial leakage control, full NS improves architecture-averaged \MacroF{} on Botswana (+8.82 pp), Indian Pines (+5.49 pp), and Pavia University (+1.59 pp), but decreases it on Salinas (-0.62 pp). The regularization-only ablation preserves most of the positive effect, whereas prototype fusion contributes only small, dataset-dependent changes and can be harmful. Stratified evaluation confirms the heterogeneity and yields corrected evidence of a full-NS degradation on Salinas and a fusion penalty on Pavia University.

The dedicated label-scarcity study further shows that the NS gain is positive on average across all tested budgets but is not monotonic in the number of labels. Although 22 of 32 fold-level slopes are directionally consistent with a larger advantage at lower $K$, none of the architecture-wise slope tests remains significant after Holm correction. The principal conclusion is therefore mechanistic rather than competitive: prototype-rule regularization can complement the structural bias of Rank-R learning, primarily by shaping the learned representation, but its utility is conditioned by dataset, architecture, and evaluation regime. Future work should test multi-prototype or structured expert-grounded rules and evaluate them with more independent spatial repetitions and neural seeds.

\section*{Acknowledgment}
Open access publication of this article was funded by HEAL-Link under an agreement between Elsevier and HEAL-Link.


\bibliographystyle{elsarticle-num}
\bibliography{references}

\end{document}